\documentclass{article}
\usepackage[utf8]{inputenc}
\usepackage{amsmath}
\usepackage{amssymb}
\usepackage{esvect}
\usepackage{float}

\usepackage{graphicx}
\usepackage{caption}
\usepackage{svg}
\graphicspath{ {./images/} }
\usepackage{authblk}

\usepackage[backend=biber, citestyle=numeric, bibstyle=numeric]{biblatex}
\usepackage[citecolor=black, urlcolor=blue, bookmarks=false, hypertexnames=true]{hyperref}

\title{Simultaneous Latent State Estimation and Latent Linear Dynamics Discovery from Image Observations}
\author[1]{Nikita Kostin}
\affil[1]{Aalto University, Finland}
\date{June 2022}

\begin{document}

\maketitle

\begin{abstract}
\noindent The problem of state estimation has a long history with many successful algorithms that allow analytical derivation or approximation of posterior filtering distribution given the noisy observations. This report tries to conclude previous works to resolve the problem of latent state estimation given image-based observations and also suggests a new solution to this problem.
\end{abstract}

\section{Introduction}
Estimating the state of the dynamical system that can't be observed directly is a very common case in many natural sciences, like physics \cite{physics} or biology \cite{biology}. To deal with that kind of problem, filtering algorithms are usually used \cite{kf}. The most general setup for filtering problems is usually written as the following:
\begin{equation}
    \begin{aligned}
        \mathbf{x}_t &= \mathbf{f}(\mathbf{x}_{t-1}, \mathbf{u}_{t-1}) \\
        \mathbf{y}_t &= \mathbf{g}(\mathbf{x}_t)
    \end{aligned}
    \label{eq:1}
\end{equation}
where dynamics function $\mathbf{f}(\cdot)$ and observation function $\mathbf{g}(\cdot)$ are stochastic non-linear functions. The mathematical form and the structure of this function highly depend on the task where state $\mathbf{x}_t$ should be estimated. The form of these functions that is particularly interesting for control theory purposes implies linear dynamics function and non-linear observation function. The dynamical system can then be written as:

\begin{equation}
    \begin{aligned}
        \mathbf{x}_t &= \mathbf{A}_{t-1} \mathbf{x}_{t-1} + \mathbf{B}_{t-1} \mathbf{u}_{t-1} + \mathbf{q_{t-1}} \\
        \mathbf{y}_t &= \mathbf{g}(\mathbf{x}_t)
    \end{aligned}
    \label{eq:2}
\end{equation}
where $\mathbf{q_{t-1}} \sim \mathcal{N}(\mathbf{q_{t-1}} | \mathbf{0}, \mathbf{Q_{t-1}})$ is zero-mean Gaussian noise and $\mathbf{g}(\cdot)$ is stochastic non-linear function that represents observation model. \\

Such a representation is preferable because linear systems are much easier to analyze compared to non-linear ones. Moreover, in control theory there are many theoretical guarantees for the stability of linear systems, see e.g. \cite{dorf_bishop}. \\

Next, different choices of observation model $\mathbf{g}(\cdot)$ structure can be made. In modern robotics practice, it's often the case that observations a robot can obtain from the environment are very high-dimensional, e.g. images. And one way of introducing such a feature in our dynamical system \eqref{eq:2} is to assume that observation model $\mathbf{g}(\cdot)$ produces high-dimensional observations $\mathbf{y}_t$ given true state $\mathbf{x}_t$. \\

The introduced dynamical model can, therefore, be used in situations, when one may suspect that a dynamical object is governed by close-to-linear dynamical law, but only high-dimensional observations of the behavior of the object can be accessed. In this case, we can approximate true dynamics by a linear function and apply a filtering algorithm to recover the true state posterior distribution given the high-dimensional observations. In this work, we are interested in scenarios when these high-dimensional observations are images.

\section{Materials and Methods}
In this section, we review components of the filtering algorithm to be applicable to the introduced system as well as describe new ideas of how limitations on these components can be overcome.
\subsection{Filtering algorithms}
In this section, it's assumed that there is no control input $\mathbf{u}_t$ yet. It's relatively easy to extend the described algorithm to also take into account the coming controls. In the next section, when introducing a new algorithm, the control inputs will be used as necessary nodes in the system architecture.
\subsubsection{Kalman Filter}
First of all, when applying the filtering algorithm the system \eqref{eq:2} with analytically given dynamical and observation models the filtering distribution $p(s_t | y_{1:t})$ can be derived using a general Bayesian filter. To show it, let's denote dynamics model probability distribution as $p(\mathbf{x}_t | \mathbf{x}_{t-1}) = \mathcal{N}(\mathbf{x}_t | \mathbf{A}_t \mathbf{x}_{t-1}, \mathbf{Q}_{t-1})$, and observation model as $\mathbf{g}(\mathbf{x}_t) = p(\mathbf{y}_t | \mathbf{x}_t)$. Then, the filtering distribution can be written as incorporating a new observation $\mathbf{y}_t$ into the previously computed $p(\mathbf{x}_t | \mathbf{y}_{1:t-1}$):
\begin{equation}
    p\left(\mathbf{x}_{t} \mid \mathbf{y}_{1:t}\right) = \frac{ p\left(\mathbf{y}_{t} \mid \mathbf{x}_{t}\right) p\left(\mathbf{x}_{t} \mid \mathbf{y}_{1:t-1}\right)}{\int p\left(\mathbf{y}_{t} \mid \mathbf{x}_{t}\right) p\left(\mathbf{x}_{t} \mid \mathbf{y}_{1:t-1}\right) \mathrm{d} \mathbf{x}_t}
\end{equation}
where $p(\mathbf{x}_t | \mathbf{y}_{1:t-1})$ can be obtained from Chapman-Kolmogorov equation:
\begin{equation}
    p\left(\mathbf{x}_{t} \mid \mathbf{y}_{1:t-1}\right) = \int p\left(\mathbf{x}_{t} \mid \mathbf{x}_{t-1}\right) p\left(\mathbf{x}_{t-1} \mid \mathbf{y}_{1:t-1}\right) \mathrm{d} \mathbf{x}_{k-1}
\end{equation}

These integrals can be computed analytically only when $\mathbf{g}(\mathbf{x}_t) = p(\mathbf{y}_t | \mathbf{x}_t)$ is a linear function of $\mathbf{x}_t$ using, e.g. Extended Kalman Filter (EKF) \cite{simo}, which uses Taylor series to linearize function $\mathbf{g}(\cdot)$. But this approach implies that there is an analytical form of observation model way to calculate derivatives $\mathbf{g}_x(\cdot)$, which is a limitation. \\

\subsubsection{Particle Filter (PF)}
Filtering distribution can also be approximated by particle filter \cite{simo} in the following way:
\begin{equation}
    p\left(\mathbf{x}_{t} \mid \mathbf{y}_{1:t}\right) \approx \sum_{i=1}^{N} w_{t}^{(i)} \delta\left(\mathbf{x}_t-\mathbf{x}_{t}^{(i)}\right),
\end{equation}
where $w_{t}^{(i)}$ is the weight of the i-th particle, and $\mathbf{x}_{t}^{(i)}$ is the state value of this i-th particle at time step $t$. Weights $w_{t}^{(i)}$ obeys the following update rule:
\begin{equation}
    w_{t}^{(i)} \propto \frac{p\left(\mathbf{y}_{t} \mid \mathbf{x}_{t}^{(i)}\right) p\left(\mathbf{x}_{t}^{(i)} \mid \mathbf{x}_{t-1}^{(i)}\right)}{\pi\left(\mathbf{x}_{t}^{(i)} \mid \mathbf{x}_{0:t-1}^{(i)}, \mathbf{y}_{1:t}\right)} w_{t-1}^{(i)},
    \label{eq:3}
\end{equation}
where $\pi\left(\mathbf{x}_{t}^{(i)} \mid \mathbf{x}_{0:t-1}^{(i)}, \mathbf{y}_{1:t}\right)$ is proposal distribution, that can be set to be a dynamics model $\pi\left(\mathbf{x}_{t}^{(i)} \mid \mathbf{x}_{0:t-1}^{(i)}, \mathbf{y}_{1:t}\right) = p\left(\mathbf{x}_{t}^{(i)} \mid \mathbf{x}_{t-1}^{(i)}\right)$, which results in Bootstrap PF weight update rule:
\begin{equation}
    w_{t}^{(i)} \propto p\left(\mathbf{y}_{t} \mid \mathbf{x}_{t}^{(i)}\right) w_{t-1}^{(i)}
\end{equation}

\subsection{Autoencoders (AE)}
However, in real-world applications, many tasks lack of analytical form of observation model $\mathbf{g}(\cdot)$ as well as dynamics matrix $\mathbf{A_t}$. Moreover, it's very often the case, when there is no clear way of how to define state space for $\mathbf{x_t}$. In those situations, such powerful function approximators like neural networks can be used to approximate $\mathbf{g}(\cdot)$ and $\mathbf{A_t}$.  \\

\begin{figure}[h] 
	\centering
	\captionsetup{width=.7\linewidth}
    \includegraphics[width=0.6\columnwidth]{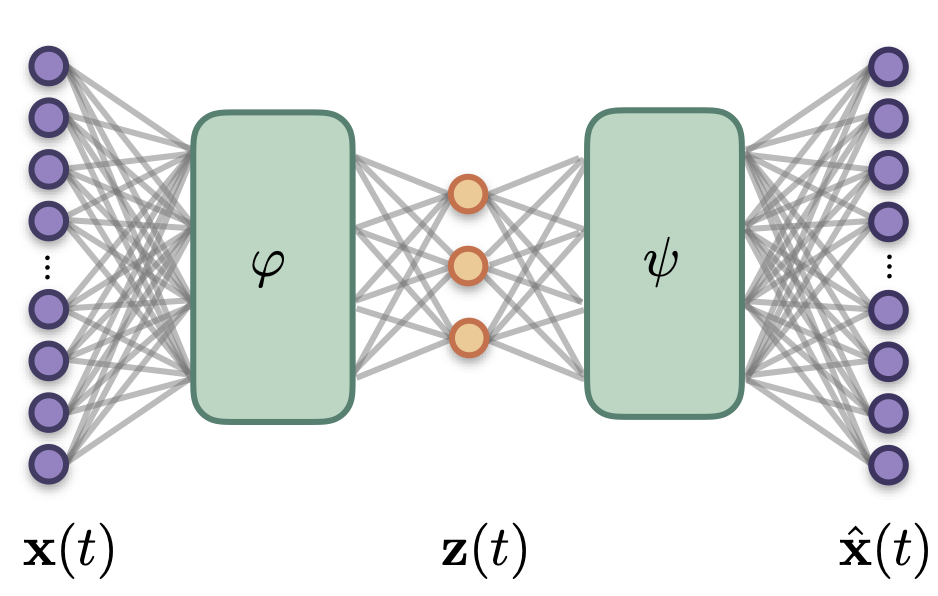}
	\caption{Autoencoder architecture: here $\mathbf{x}(t)$ stands for high-dimensional observation, $\mathbf{z}(t)$ stands for latent state and $\mathbf{\hat{x}}(t)$ is reconstructed observation by the decoder $\psi$.}
	\label{fig:1}
\end{figure}

One specific architecture of neural network called Autoencoder, which is a feed-forward neural network that consists of two parts: encoder, which encodes observation from high-dimensional space to low-dimensional latent space, and decoder, which tries to reconstruct original observation, based only on its low-dimensional representation given by encoder (Fig.\ref{fig:1}).

The way to use AE architecture for approximating the observation model is straightforward: decoder $\psi$ plays the role of the observation model. Training of AE requires only observations $\{\mathbf{y}_t\}_{t=1}^N$. The only question is how can we formulate latent state $\mathbf{x_t}$ (on Fig.\ref{fig:1} it's denoted as $\mathbf{z}(t)$) that each of observations $\mathbf{y_t}$ corresponds to? \\

One could approach this problem using, for example, Sparse Identification of Nonlinear Dynamics (SINDy) \cite{sindy, ae_sindy}, which makes the use of latent state polynomials permutation matrix and applies sparsity-promoting linear regression to it, which results in approximate, but analytical dynamics expression. And SINDy \cite{ae_sindy} does it using AE as a function to transit between observation and latent space. During training, SINDy learns both AE parameters and latent dynamics representation simultaneously. \\
However, this method can't be used to compute probability densities of dynamics and observation models, which are necessary if, for example, a particle filter is used.

\subsection{Normalizing Flows (NF)}
Normalizing Flows \cite{nf_review} is a family of methods that allow to compute an exact density of $\mathbf{y}$ sampled from some distribution $p(\mathbf{y})$. It's possible since the change-of-variables formula can be written as
\begin{equation}
    p_{\mathrm{y}}(\mathbf{y}) = p_{\mathrm{u}}(\mathbf{u}) \left | \operatorname{det} J_{T}(\mathbf{u})\right|^{-1} \quad \text { where } \quad \mathbf{u} = T^{-1}(\mathbf{y}).
    \label{eq:4}
\end{equation}

Here $\left | \operatorname{det} J_{T}(\mathbf{u})\right|$ is Jacobian of the flow $T$ from $\mathbf{y}$ to $\mathbf{u}$. Base distribution $p_{\mathrm{u}}(\mathbf{u})$ is usually chosen to be a simple-to-sample and simple-to-evaluate distribution, like multivariate Gaussian with some mean and covariance. \\

This approach can be merged with PF, where several distribution approximations require evaluation of its densities. First, NF can be used to model dynamics function, if it's not required for it to be linear \cite{cond_nf_pf}. \\

Second, the proposal distribution that is used for sampling particles in \eqref{eq:3}, can also be modeled using NF (known as particle filter using conditional NF \cite{cond_nf_pf}), since it allows not only to compute exact density $\pi\left(\mathbf{x}_{t}^{(i)} \mid \mathbf{x}_{0:t-1}^{(i)}, \mathbf{y}_{1:t}\right)$, but also allows to sample from this distribution using samples from the much simpler distribution $p_{\mathrm{u}}(\mathbf{u})$. \\

Finally, NF can model observation function $\mathbf{g}(\cdot)$ as $\mathbf{g}_{\theta}(\cdot)$, because NF allow to evaluate observation likelihood $p(\mathbf{y} | \mathbf{x})$. As a theoretical advantage of this approach, it would allow us to avoid keeping both encoder and decoder as in AE because NF can be inverted, so it is able to encode the observation into a latent state and decode the latent state into observation. \\

But straightforward implementation of observation model as NF encounters the key property of conventional NF: the dimensionalities of $\mathbf{y}$ and $\mathbf{u}$ should be the same to ensure invertibility. There are new ways to formulate NF in a way to allow flow dimensionality reduction, e.g. \cite{nf_across_dim, funnel}, and it should be investigated. But in this work, we follow a different approach.

\section{Normalizing Flows based Particle Filter (NFPF)}
Several algorithms from the previous section, namely SINDy and particle filter using conditional NF, can be used in the setting that was described in the introduction to enable results comparison with the method that we describe in this section. \\

\subsection{Modelling observation likelihood using conditional NF}
\begin{figure}[t] 
	\raggedleft
    \includegraphics[width=\linewidth]{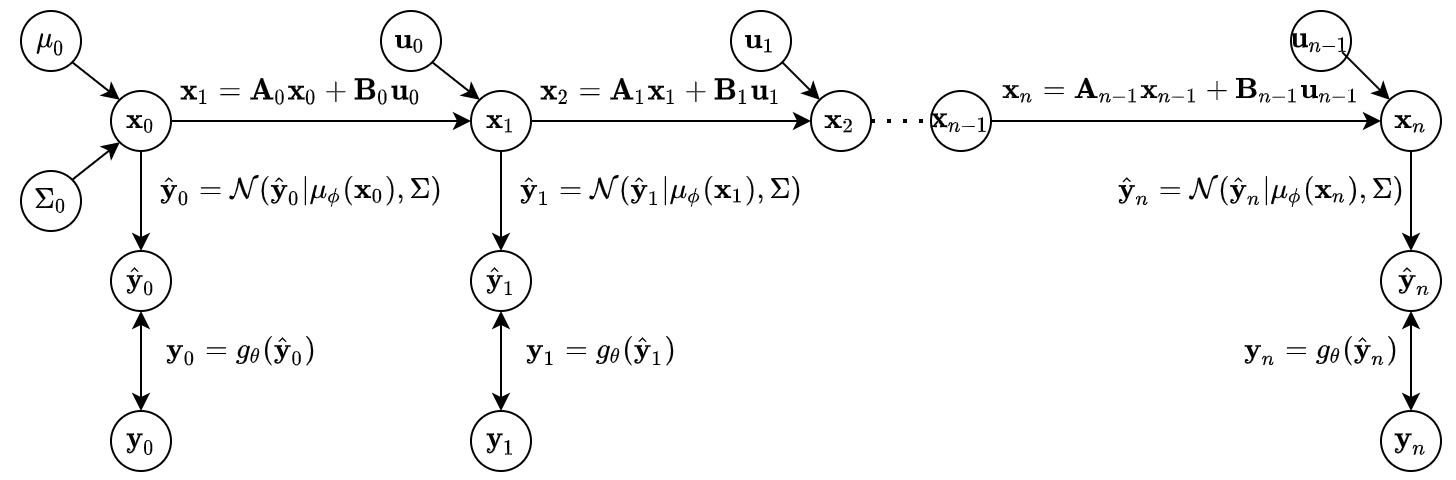}
	\caption{Proposed model architecture: $\mathbf{y}_t$ denotes image observation, $\hat{\mathbf{y}}_t$ denotes flow base distributed variables (Gaussian), $\mathbf{x}_t$ denotes latent state, $\mathbf{A}_t$ denotes dynamics matrix, $\mathbf{u}_t$ denotes control signal, $\mathbf{B}_t$ denotes control matrix, $\mu_0$ and $\Sigma_0$ are mean and covariance of the first latent state. Linear dynamics matrices are obtained using dynamics network $f_{\psi}$, $g_{\theta}$ denotes NF transformation, while $\mu_{\phi}$ stands for mean parametrization for $\hat{\mathbf{y}}$}.
	\label{fig:2}
\end{figure}

To be able to estimate an exact density $p(\mathbf{y} | \mathbf{x})$, we employ conditional NF \cite{cond_nf} to operate a Hidden Markov Model (HMM), shown in Fig.\ref{fig:2}, and model observation density in terms of simpler base distribution density.

In this model, we make use of conditional NF \cite{cond_nf} that preserve the dimensionality to model a complex observation distribution $p(\mathbf{y} | \mathbf{x})$ using samples from a simple Gaussian distribution $p(\hat{\mathbf{y}}_t | \mathbf{x}_t)$ for which we parameterize the mean of the Gaussian distribution $p(\hat{\mathbf{y}}_t | \mathbf{x}_t) = \mathcal{N}(\hat{\mathbf{y}}_t | \mu, \Sigma)$ by applying convolutional neural network to latent state $\mathbf{x}_t$ to produce $\mu = \mu_{\phi}(\mathbf{x}_t)$ and using constant covariance $\Sigma$. \\

As a result, this method allows joint training of NF $g_{\theta}$, mean parametrization $\mu_{\phi}$ and linear system matrices $\mathbf{A}_t$ and $\mathbf{B}_t$  jointly, since the whole model is differentiable and the gradients can flow through each arrow. Such a system also makes it possible to obtain state-dependent dynamics and control matrices using an additional neural network with parameters $\psi$ that takes $\mathbf{x}_t$ as input and outputs $\mathbf{A}_t$ and $\mathbf{B}_t$. It's even possible to incorporate control signal $\mathbf{u}_t$ as an extra input into this network to allow more flexibility in linear dynamics matrices, but in our experiments, we avoided this approach for now. The training objective is maximizing the log-likelihood of the data taken along a batch of size $K$ over latent states and observations starting of time step $t=k$:
\begin{equation}
    \begin{aligned}
        \max_{\theta, \phi, \psi} \log p(Y | X)  &= \max_{\theta, \phi, \psi} \sum_{t=k}^{k + K} \log p(\mathbf{y}_t | \mathbf{x}_t) = \\ 
        &= \max_{\theta, \phi, \psi} \sum_{t=k}^{k + K} \log \left( p(\hat{\mathbf{y}}_t | \mu_{\phi}(\mathbf{x}_t), \Sigma) \left | \operatorname{det} J_{g_{\theta}^{-1}}(\mathbf{y}_t) \right| \right)
    \end{aligned}
    \label{eq:5}
\end{equation}

If one would recall expression for NF density \eqref{eq:4}, then inner of optimizing in \eqref{eq:5} objective can be expanded as the following:
\begin{equation}
    \begin{aligned}
        \log p(Y | X) &= \sum_{t=k}^{k + K} \log \left( p(\hat{\mathbf{y}}_t | \mu_{\phi}(\mathbf{A}_{\psi} \mathbf{x}_{t-1} + \mathbf{B}_{\psi} \mathbf{u}_{t-1}), \Sigma) \left | \operatorname{det} J_{g_{\theta}^{-1}}(\mathbf{y}_t) \right| \right) = \\
        &= \sum_{t=k}^{k + K} \log p(\hat{\mathbf{y}}_t | \mu_{\phi}(\mathbf{A}_{\psi} \mathbf{x}_{t-1} + \mathbf{B}_{\psi} \mathbf{u}_{t-1}), \Sigma) + \log  \left | \operatorname{det} J_{g_{\theta}^{-1}}(\mathbf{y}_t) \right|
    \end{aligned}
    \label{eq:6}
\end{equation}

As it can be seen from \eqref{eq:6}, maximization of log-likelihood, or minimization of negative log-likelihood, allows to update all the parameters in the model:
\begin{equation}
    \begin{aligned}
        \min_{\theta, \phi, \psi} - \log p(Y | X) = &\min_{\phi, \psi} \sum_{t=k}^{k + K} \log p(\hat{\mathbf{y}}_t | \mu_{\phi}(\mathbf{A}_{\psi} \mathbf{x}_{t-1} + \mathbf{B}_{\psi} \mathbf{u}_{t-1}), \Sigma) \ + \\
        &+ \min_{\theta} \sum_{t=k}^{k + K} \log  \left | \operatorname{det} J_{g_{\theta}^{-1}}(\mathbf{y}_t) \right|
    \end{aligned}
\end{equation}

Latent dynamical system matrices parametrized by neural network $f_{\psi}$ should restrict these matrices to have nice features, inherited from linear systems theory, e.g. spectral radius of $\mathbf{A}_{\psi}$ should be less than 1 to ensure asymptotic stability of the system \cite{lin_sys_stab}. Also the controllability property for matrix pair $\mathbf{A}_{\psi}$ and $\mathbf{B}_{\psi}$ can be imposed. \\

In the end, the algorithm allows to evaluate observation likelihood $p(\mathbf{y}_{t} \mid \mathbf{x}_{t})$, which can be then used in particle filter to compute weights of the particles \eqref{eq:3}.

\section{Experimental evaluation}
\begin{figure}[H]
    \centering
    \includegraphics[width=0.5\linewidth]{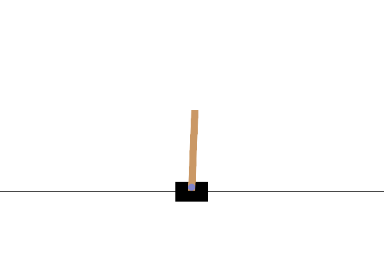}
	\caption{RGB observation of CartPole used in the experiments.}
	\label{fig:3}
\end{figure}

NFPF algorithm described above is implemented using Python language within PyTorch Deep Learning framework \cite{torch}. As an example of the dynamical model the CartPole from OpenAI Gym classic control framework \cite{brockman2016openai} which is a pole that needs to be stabilized on the moving cart. The typical observation for this environment is shown in Fig.\ref{fig:3}. \\

The training setup begins with training a simple agent that can provide trajectories long enough to train the algorithm. As such an agent the DQN algorithm \cite{dqn} was chosen. The data collected by the DQN agent was then used to train the NF filtering algorithm using the scheme from Fig.\ref{fig:2}. The dynamics matrices parametrization $f_{\psi}$ was implemented to output matrices with unit Frobenius norm, which should, in principle, guarantee that these matrices' spectral radius is less than 1. The dimensionality of the latent space was set to be the same number as the CartPole state space dimensionality to 4. \\

\begin{figure}[H]
    \centering
    \includegraphics[width=1.\linewidth]{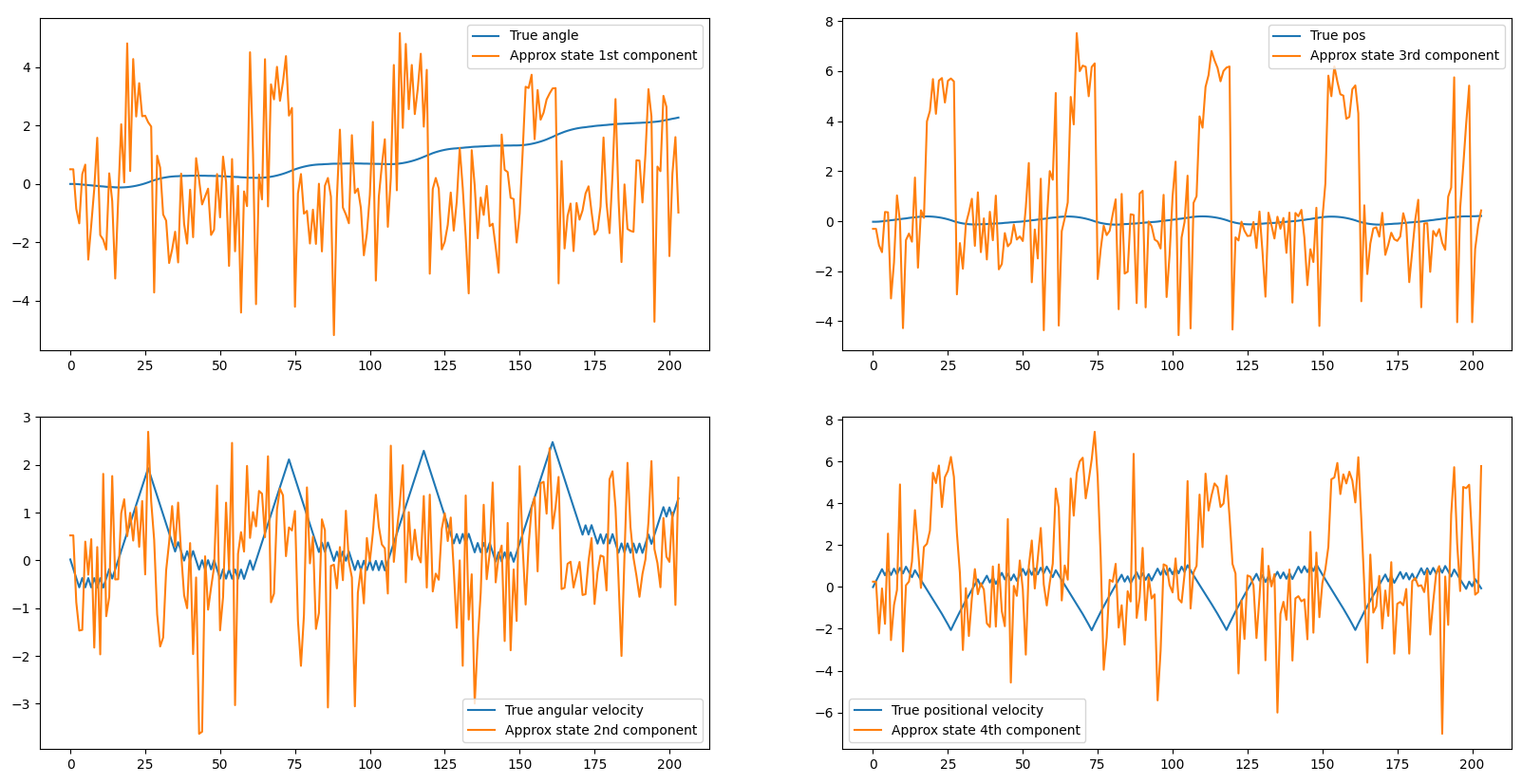}
	\caption{Particle filter with 2 particles NF observation likelihood against true state evolution given the same control inputs for 4 dimensions.}
	\label{fig:4}
\end{figure}

\begin{figure}[H]
    \centering
    \includegraphics[width=1.\linewidth]{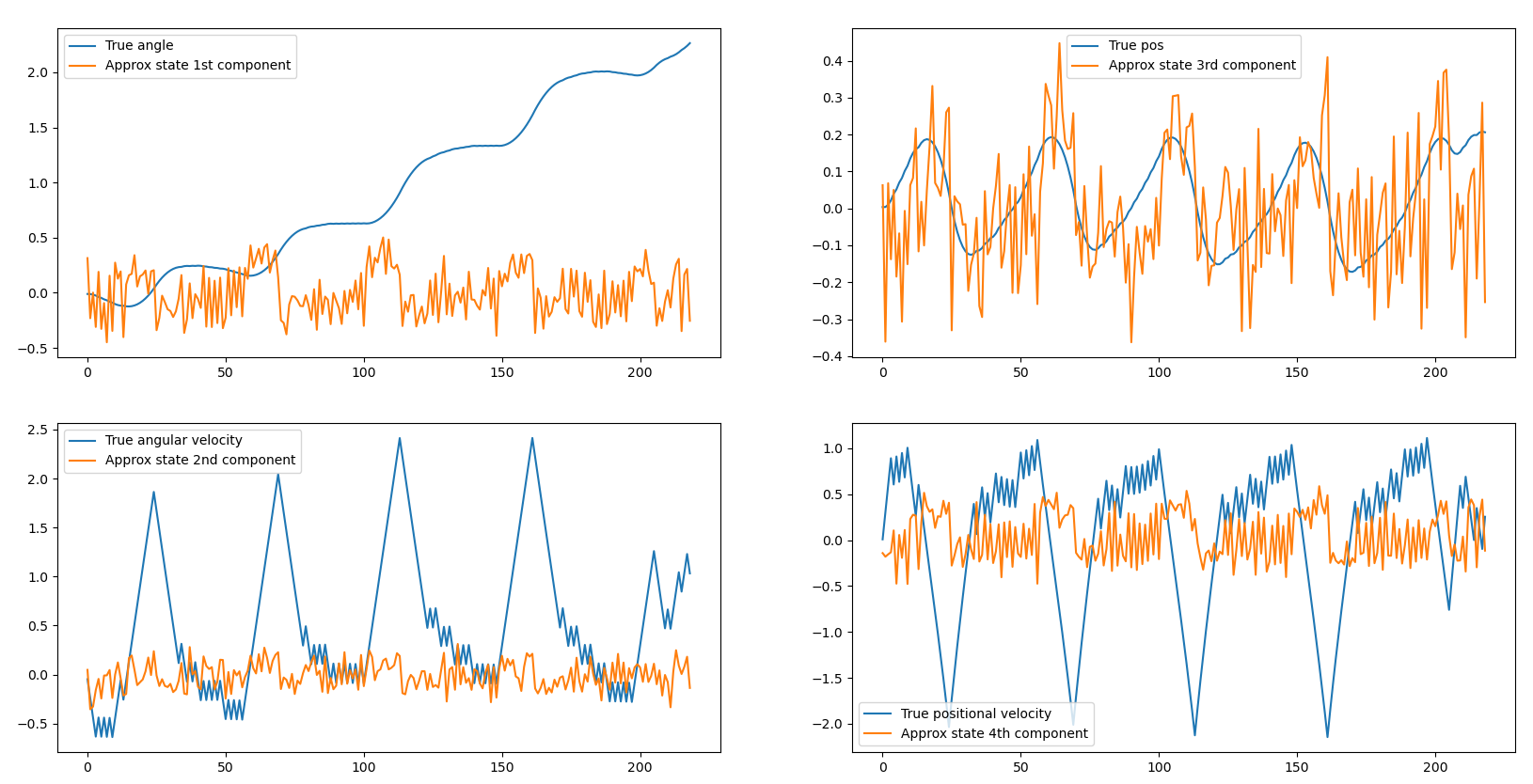}
	\caption{PF with 2 particles NF observation likelihood against true state evolution given the same control inputs for 100 dimensions.}
	\label{fig:5}
\end{figure}

To check what kind of latent states were produced by the described algorithm, the bootstrap particle filter takes NF observation likelihood and standard normal distributed latent state and outputs the next step latent states delta-functions posterior distribution. The resulting plots for one trajectory with mean among the particles are shown in Fig.\ref{fig:4} and Fig.\ref{fig:5}.

\section{Discussion and Conclusion}
The resulting plots from the previous section might not convince the reader that the proposed algorithm does produce reasonable latent state representation, but there is a lack of experiments, because:
\begin{enumerate}
    \item Only 10 trajectories were used to train NF observation likelihood since it requires a lot of computing power to train NF.
    \item More restrictions on dynamics matrices should be imposed, e.g. controllability restriction, since it's an important one and it wasn't used in the provided experiments.
    \item Experiments testing the richness of the latent state for the control algorithms should be conducted, e.g. LQR in the latent space.
    \item Only a bootstrap particle filter was used in the experiments, while good proposal distribution for generating particles can be incorporated into the algorithm to enhance its performance.
\end{enumerate}

Summarizing the report, the problem of state estimation given image observation can be addressed by several approaches. Each of them has its advantages and disadvantages depending on the particular task formulation. In the lack of dynamics and observation models, only a few algorithms can produce the posterior filtering distribution. \\

SINDy employs autoencoder architecture to approximate the observation model and allows simultaneous training of the dynamics model. The method was proved to work empirically. However, it requires training both the encoder and the decoder while only the decoder is needed to estimate the observation model. On the other hand, the encoder can be used to encode image observations to map them into a latent state. \\

Observation model estimation using NFPF provides a clear way of obtaining observation likelihood, while also producing time-dependent matrices of linear dynamics. Observation likelihood allows to obtain the filtering distribution using the particle filter. A disadvantage of this approach is that NF requires a lot of parameters to achieve enough expressivity of the model. Another drawback of this approach is the absence of a batch training algorithm: there is no easy way to train NFPF using batches of observation rather than single images. It means very slow convergence of the algorithm as well as very noisy gradients. \\

In conclusion, SINDy showed that observation data is enough to infer the latent state dynamical model and provides an efficient way of discovering latent state dynamics. Conditional NF observation likelihood estimation, on the other hand, provides a full posterior filtering distribution, rather than just a sample from this distribution, but it requires fixing a lot of issues and very careful training.

\newpage
\printbibliography

\end{document}